\documentclass[10pt,twocolumn,letterpaper]{article}

\usepackage{CVPR2024/cvpr}       % To produce the CAMERA-READY version
\usepackage[accsupp]{axessibility}
\usepackage[dvipsnames]{xcolor}

\definecolor{cvprblue}{rgb}{0.21,0.49,0.74}
\usepackage[pagebackref,breaklinks,colorlinks,citecolor=cvprblue]{hyperref}

\def\paperID{66} % *** Enter the Paper ID here
\def\confName{CVPR}
\def\confYear{2024}

\title{Recursive Joint Cross-Modal Attention for Multimodal Fusion in Dimensional Emotion Recognition}

\author{R. Gnana Praveen ~~~~ Jahangir Alam \\ 
Computer Research Institute of Montreal (CRIM), Canada\\
{\tt\small {gnana-praveen.rajasekhar@crim.ca ~ jahangir.alam}@crim.ca}
}

\begin{document}
\maketitle
\begin{abstract}

%Multimodal Emotion Recognition (MER) is found to be promising as they can outperform unimodal approaches by leveraging the diverse and complementary relationships across multiple modalities. 
Though multimodal emotion recognition has achieved significant progress over recent years, %has been achieved using current Multimodal Emotion Recognition approaches, 
the potential of rich synergic relationships across the modalities is not fully exploited. %they still suffer from heterogeneity gaps across modalities. 
In this paper, we introduce Recursive Joint Cross-Modal Attention (RJCMA) to effectively capture both intra- and inter-modal relationships across audio, visual, and text modalities for dimensional emotion recognition. In particular, we compute the attention weights based on cross-correlation between the joint audio-visual-text feature representations and the feature representations of individual modalities to simultaneously capture intra- and inter-modal relationships across the modalities. The attended features of the individual modalities are again fed as input to the fusion model in a recursive mechanism to obtain more refined feature representations. We have also explored Temporal Convolutional Networks (TCNs) to improve the temporal modeling of the feature representations of individual modalities. Extensive experiments %\footnote{The code is available on GitHub: \url{https://github.com/praveena2j/RJCMA}} 
are conducted to evaluate the performance of the proposed fusion model on the challenging Affwild2 dataset. By effectively capturing the synergic intra- and inter-modal relationships across audio, visual, and text modalities, the proposed fusion model achieves %second place in the valence arousal challenge of 6th Affective Behavior Analysis in-the-Wild (ABAW) competition with 
a Concordance Correlation Coefficient (CCC) of 0.585 (0.542) and 0.674 (0.619) for valence and arousal respectively on the validation set (test set). This shows a significant improvement over the baseline of 0.240 (0.211) and 0.200 (0.191) for valence and arousal, respectively, in the validation set (test set), achieving second place in the valence-arousal challenge of the 6th Affective Behavior Analysis in-the-Wild (ABAW) competition. 
The code is available on GitHub: \url{https://github.com/praveena2j/RJCMA}.
\end{abstract}    
\section{Introduction}
\label{sec:intro}
%%% why multimodal emotion recognition
Emotion recognition is a challenging problem due to the diverse nature of expressions associated with emotional states across individuals and cultures \cite{9472932}. It has a wide range of applications in various fields such as health care (assessing pain, fatigue, depression) \cite{9320216,RAJASEKHAR2021104167}, autonomous driving (assessing driver emotional states) \cite{CHEN2022104569}, robotics (for realistic human-computer interaction) \cite{10.1145/1101149.1101299}, etc. Emotion recognition is often explored in the literature as a classification problem of categorizing emotions into seven classes: anger, disgust, fear, happiness, sadness, surprise, and contempt \cite{Ekman}. Recently, compound expression datasets have also been introduced to capture human emotions beyond seven classes \cite{kollias2023multi}. However, they also do not capture the diverse range of emotions expressed by humans. Therefore, a dimensional model of emotions has been introduced to capture the diverse range of emotions, predominantly using valence and arousal. % as shown in Fig \ref{fig:VA}. 
Valence denotes the range of emotions from very sad (negative) to very happy (positive), whereas arousal represents the intensities of emotions from being very passive (sleepiness) to very active (high excitement) \cite{Schlosberg}. Dimensional Emotion Recognition (DER) is more challenging than categorical emotion recognition, as they are highly prone to label noise due to the complex process of obtaining dimensional annotations.        
%\begin{figure}[!t]
%\centering
%\includegraphics[width=0.20\textwidth]{sec/Figures/ValenceArousal.png}
%\caption{The valence-arousal space %\cite{Praveen_2022_CVPR}}
%\label{fig:VA}
%\end{figure}

Multimodal learning has been recently gaining a lot of attention as it can offer rich complementary information across multiple modalities, which can play a crucial role in outperforming unimodal approaches \cite{9758834}. Human emotions are communicated in complex ways through various modalities such as face, voice, and language. Multimodal emotion recognition aims to effectively fuse audio, visual, and text by capturing the rich intra- and inter-modal complementary relationships across the modalities. Early approaches of multimodal fusion for DER either rely on Long Short Term Memory (LSTM)-based fusion \cite{cite7,cite6} or early feature concatenation \cite{9320301,8914655}. %With the advancement of sophisticated deep learning models \cite{}, the performance of MER has been significantly improved. 
With the advent of transformers \cite{NIPS2017_3f5ee243}, attention models using multimodal transformers have attained much interest to effectively combine multiple modalities for DER \cite{srini_2021_SLT,Situ-RUCAIM3,10208807}. Recently, cross-modal attention was found to be quite promising in capturing the rich inter-modal complementary relationships across the modalities, which has been successfully applied for several applications such as action localization \cite{lee2021crossattentional}, emotion recognition \cite{9667055}, and person verification \cite{10.1007/978-3-031-48312-7_2}. Praveen et al. \cite{Praveen_2022_CVPR,10005783} explored joint cross-attention by introducing joint feature representation in the cross-attentional framework and achieved significant improvement over vanilla cross-attention \cite{9667055}. They have further improved the performance of the model by introducing recursive fusion, and LSTMs for temporal modeling of individual feature representations as well as the audio-visual representations \cite{10095234}. Recursive fusion of the cross-attention models has also been successfully explored %was also found to be promising 
for other audio-visual tasks such as event localization \cite{9423042} and person verification \cite{praveen2024audio}. However, most of the existing cross-attention based models are focused on audio-visual fusion for DER. 

In this work, we have investigated the prospect of efficiently capturing the synergic intra- and inter-modal relationships across audio, visual, and text modalities for DER. By deploying cross-correlation between the joint audio-visual-text feature representation and feature representations of individual modalities, %and joint representation of the other two modalities
we can simultaneously capture intra- and inter-modal relationships across the modalities. Inspired by the performance of recursive attention models \cite{9423042,10095234}, we have also incorporated a recursive fusion of audio, visual, and text modalities in the context of joint cross-attentional fusion to obtain more refined feature representations. 
%The attended features of the individual modalities are further refined using the recursive mechanism to obtain robust feature representations. 
The major contributions of the paper can be summarized as follows.

\begin{itemize}
    \item Joint cross-modal attention is explored among audio, visual, and text modalities using a joint audio-visual-text feature representation to simultaneously capture intra- and inter-modal relationships across the modalities.
    \item Recursive fusion is used to further improve the feature representations of individual modalities. TCNs are also used to improve the temporal modeling of individual feature representations. 
    \item Extensive experiments are conducted to evaluate the robustness of the proposed approach on the challenging Affwild2 dataset. 
\end{itemize}
%Recursive fusion has been previously applied successfully for various audio-visual tasks such as event localization, person verification and emotion recognition in the frame work of joint cross-attention. In this work, we have explored recursive fusion in the context of cross-modal attention across audio, visual, and text modalities for dimsnionsl emotion recognition

\section{Related Work}
\subsection{Multimodal Emotion Recognition}

One of the early approaches using deep learning architectures for DER was proposed by Tzirakis et al. \cite{cite7}, where they explored 1D Convolutional Neural Networks (CNN) for audio and Resnet-50 \cite{7780459} for visual modality. The deep features are then fused using LSTMs for estimating predictions of valence and arousal. With the advancement of 3D CNN models, Kuhnke et al. \cite{9320301} showed performance improvement using R3D \cite{8578773} for visual modality and Resnet-18 \cite{7780459} for audio modality with simple feature concatenation. Kollias et al. \cite{kollias2021affect,kollias2021distribution,kollias2019face} explored DER along with other tasks of classification of expressions and action units in a unified framework. Another widely explored line of research for DER is based on knowledge distillation (KD) \cite{hinton2015distilling}. Schoneveld et al. \cite{cite6} explored KD for obtaining robust visual representations, while  Wang et al. \cite{9607711} and Deng et al. \cite{Deng_2021_ICCV} attempted to leverage KD to deal with label uncertainties. Recently, KD is also explored along with the paradigm of Learning Under Privileged Information (LUPI) to efficiently exploit multiple modalities for DER \cite{10208547}. Li et al. \cite{Li_2023_CVPR} proposed Decoupled Multimodal Distillation (DMD) to mitigate the issues of multimodal heterogeneities by dynamically distilling the modality-relevant information across the modalities. Although these methods have shown promising performance by exploiting multiple modalities, they do not focus on capturing the synergic relationships pertaining to intra- and inter-modal relationships across the modalities.  

\subsection{Attention Models for Multimodal Emotion Recognition}
Inspired by the performance of transformers \cite{NIPS2017_3f5ee243}, several approaches have been proposed to investigate the potential of transformers for DER. Most of the existing works explored transformers to encode the concatenated version of the feature representations of individual modalities \cite{Situ-RUCAIM3,10208807,10209026,10208713}. Tran et al. \cite{9747278} showed that fine-tuning transformers with cross-modal attention trained on a large-scale voxceleb2 \cite{Chung18b} dataset helps to improve the performance of multimodal emotion recognition. Huang et al. \cite{9053762} explored multimodal transformers along with self-attention modules, where audio modality is used to attend to visual modality to produce robust multimodal feature representations. Parthasarathy et al. \cite{srini_2021_SLT} further extended their idea by employing cross-modal attention in a bidirectional fashion, where audio modality is used to attend to visual modality and vice-versa. Karas et al. \cite{AU-NO} provided a comprehensive evaluation of the fusion models based on self-attention, cross-attention and LSTMs for DER. Zhang et al. \cite{10208757} proposed leader-follower attention, where audio, visual, and text modalities are combined to obtain the attended feature representations using the modality-wise attention scores, which is further concatenated to visual features for final predictions. 

Praveen et al. \cite{9667055} explored cross-attention based on cross-correlation across the feature representations of audio and visual modalities. They further extended their approach by employing joint audio-visual feature representation in the cross-attentional framework and showed significant performance improvement \cite{Praveen_2022_CVPR}. Praveen et al. \cite{10095234} improved the performance by introducing recursive fusion and LSTMs for temporal modeling of individual modalities. In this work, we further extend the idea of \cite{10095234} by incorporating text modality, and TCNs to effectively capture the intra-modal relationships. The proposed approach primarily differs from \cite{10095234} in three aspects : (1) In this work, we employ text modality in the framework of recursive joint cross-attention in addition to audio and visual modalities, whereas \cite{10095234} uses only audio and visual modalities. (2) In \cite{10095234}, LSTMs are used, whereas we have deployed TCNs as they are found to be effective in improving the temporal modeling of individual modalities. (3) In \cite{10095234}, R3D \cite{8578773} is used for visual backbone and Resnet 18 for audio modality, whereas we have used Resnet-50 %pretrained on MS-CELEB-1M \cite{}, which is further 
fine-tuned on FER+ \cite{barsoum2016training} for visual and VGG \cite{hershey2017cnn} pretrained on Audioset for audio modality.     

\section{Proposed Approach}

\subsection{Visual Network}
Facial expressions play a significant role in conveying the emotional state of a person. In videos, spatial information provide the semantic regions of face pertinent to the expressions, whereas temporal dynamics convey the evolution of expressions in videos across the frames. Therefore, effectively modeling the spatial and temporal dynamics of facial expressions in videos is crucial to obtain robust visual feature representations. Several approaches have been explored using 2D CNNs in conjunction with LSTMs, where 2D CNNs are used to encode the spatial information and LSTMs for temporal dynamics of facial expressions \cite{5740839,WOLLMER2013153}. With the advent of 3D CNN models \cite{8578773}, they are successfully explored for DER \cite{9320301,Praveen_2022_CVPR,10005783} and showed improvement over 2D CNNs with LSTMs. It has also been shown that 3DCNNs in combination with LSTMs are effective in capturing the temporal dynamics \cite{10095234}, where 3DCNNs are efficient in capturing short-term dynamics and LSTMs are robust in modeling long-term dynamics. Recently, TCNs are found to be promising to effectively capture the temporal dynamics for DER \cite{10208757,10209026}. In this work, we have used Resnet-50 \cite{7780459} pretrained on MS-CELEB-1M dataset \cite{10.1007/978-3-319-46487-9_6}, which is further fine-tuned on FER+ \cite{barsoum2016training} dataset similar to that of \cite{10208757}. We have further used TCNs to effectively capture the temporal dynamics of facial expressions. 

\subsection{Audio Network}
Speech-based emotion recognition is another promising research area due to the rich emotion-relevant information in the vocal expressions. With the advancement of deep learning models, vocal expressions are encoded using 1D CNNs with raw speech signals \cite{cite7} or 2D CNNs with spectrograms \cite{cite6,9320301}. Some of the works also explored deep features in combination with conventional hand-crafted features to encode the vocal expressions \cite{Situ-RUCAIM3,9607460}. Recently, spectrograms have been widely explored as they are found to be efficient in capturing the affective states of the vocal expressions \cite{10208757,Praveen_2022_CVPR}. Therefore, we have also explored spectrograms with 2D CNNs to encode the vocal expressions. Specifically, we have used VGG-Net architecture pretrained on large-scale audioset dataset \cite{45611}. Similar to visual modality, we have also used TCNs to encode the temporal dynamics of frame-level vocal embeddings.  

\subsection{Text Network}
Text modality is another predominantly explored modality for emotion detection, which carries semantic emotion-relevant information in the text data \cite{10409495}. Effectively leveraging the textual data can boost the performance of multimodal fusion as they can offer significant emotion-relevant information and complement audio and visual modalities. Based on transformers, BERT features are predominantly explored text encoders for emotion recognition in the literature \cite{Acheampong2021TransformerMF}. Therefore, we also used BERT as text encoder, followed by TCNs to encode the temporal information across the word embeddings. % similar to that of \cite{10208757}. 

\begin{figure*}[htb]
\centering
\centerline{\includegraphics[width=0.9\linewidth]{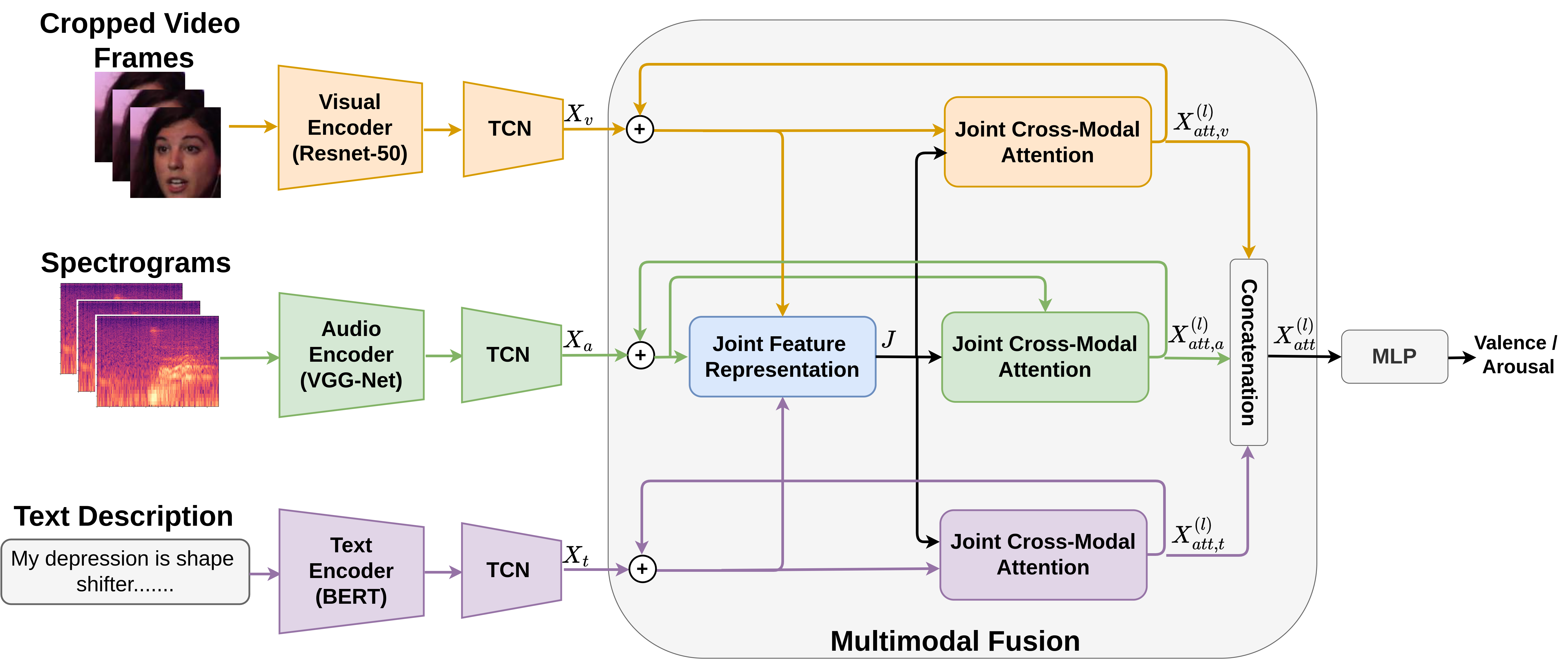}}
\caption{Illustration of the proposed framework with the recursive joint cross-modal attention. Best viewed in color.} 
\label{fig:BD}
\end{figure*}

\subsection{Recursive Joint Cross-Modal Attention}
Given the input video sub-sequence $S$ of $K$ frames, the audio, visual and text data are preprocessed and fed to the corresponding encoders, followed by TCNs to obtain the feature representations of the respective modalities as shown in Fig \ref{fig:BD}. The feature representations of the audio, visual and text modalities are denoted by ${\boldsymbol X}_{a}\boldsymbol =  \{ \boldsymbol x_{a}^1, \boldsymbol x_{a}^2, ..., \boldsymbol x_{a}^K \boldsymbol \} \in \mathbb{R}^{d_a\times K}$,
${\boldsymbol X}_{v}\boldsymbol =  \{ \boldsymbol x_{v}^1, \boldsymbol x_{v}^2, ..., \boldsymbol x_{v}^K \boldsymbol \} \in \mathbb{R}^{d_v\times K}$ and ${\boldsymbol X}_{t}\boldsymbol =  \{ \boldsymbol x_{t}^1, \boldsymbol x_{t}^2, ..., \boldsymbol x_{t}^K \boldsymbol \} \in \mathbb{R}^{d_t\times K}$ respectively, where $d_a$, $d_v$, and $d_t$ are dimensions of audio, visual and text features. $\boldsymbol x_{a}^k$, $\boldsymbol x_{v}^k$ and $\boldsymbol x_{t}^k$ represents the feature vectors of the individual frames of audio, visual, and text modalities.  

Given the audio (A), Visual (V), and Text (T) feature representations, $\boldsymbol X_a$, $\boldsymbol X_v$, and $\boldsymbol X_t$ the joint feature representation ($\boldsymbol J$) of audio, visual, and text modalities is obtained by concatenating the feature vectors of all modalities, followed by fully connected layer as 
\begin{equation}
 {\boldsymbol J = FC([{\boldsymbol X}_{a} ; {\boldsymbol X}_{v} ; {\boldsymbol X}_{t}]) \in\mathbb{R}^{d\times K}} 
\end{equation}
where $d = {d_a} + {d_v} + {d_t}$ denotes the dimensionality of $\boldsymbol J$ and $FC$ denote fully connected layer. 

\begin{figure*}[htb]
\centering
\centerline{\includegraphics[width=0.7\linewidth]{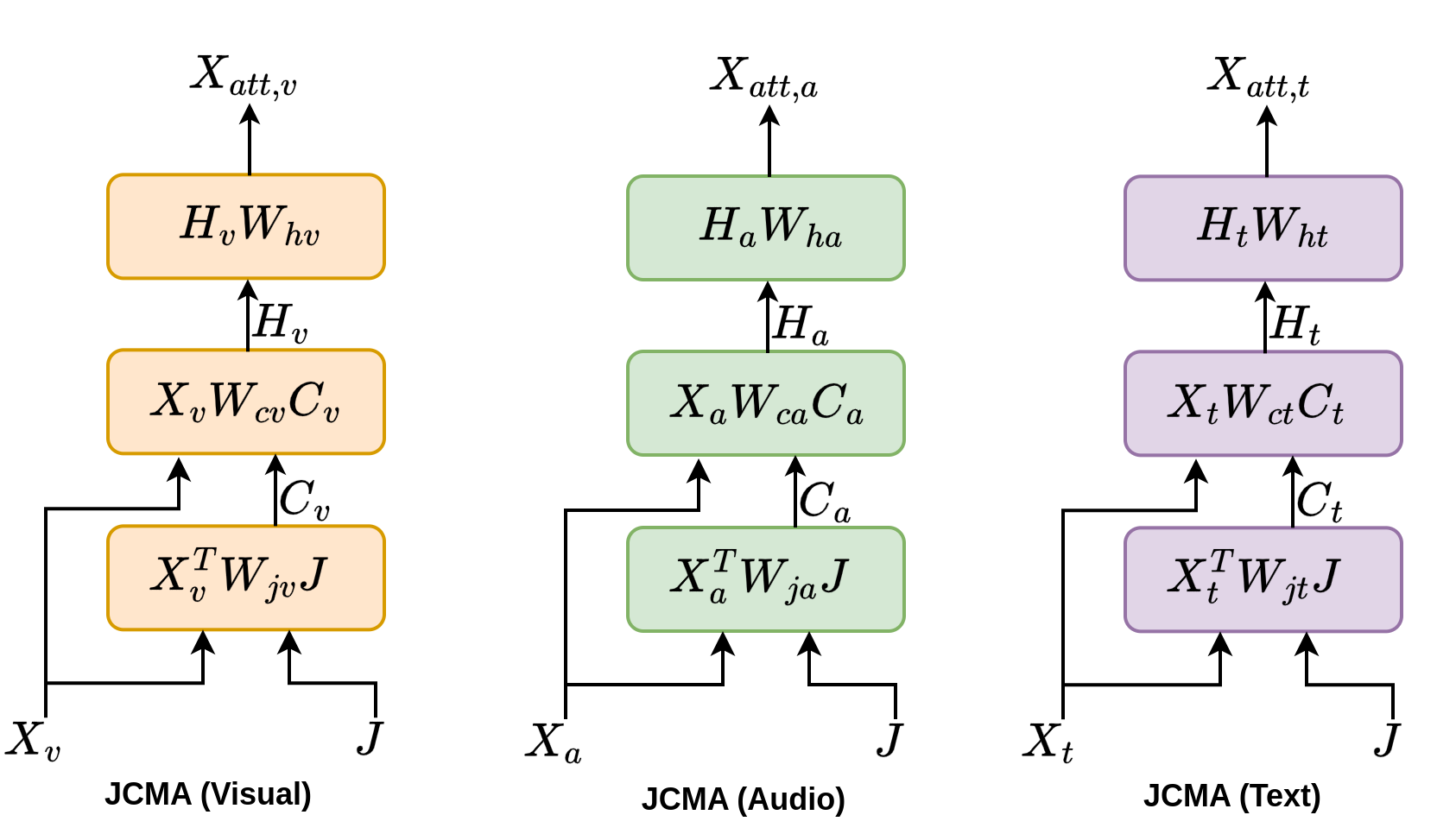}}
\caption{Joint Cross-Modal Attention blocks of audio, visual and text modalities} \label{fig:JCMA}
\end{figure*}

Now $\boldsymbol J$ is fed to the joint cross-attentional framework of the respective modalities as shown in Fig \ref{fig:JCMA} to attend to the feature representations of individual modalities. This helps to simultaneously encode the intra- and inter-modal relationships within the same modalities as well as across the modalities in obtaining the attention weights. The cross-correlation between the $\boldsymbol J$ and ${\boldsymbol X}_{a}$ are obtained as joint cross-correlation matrix $\boldsymbol C_{a}$, which is given by 

%The concatenated A-T feature representations ($\boldsymbol J$) of a video sub-sequence ($\boldsymbol S$) are now used to attend to unimodal feature representations ${\boldsymbol X}_{\mathbf a}$ and ${\boldsymbol X}_{\mathbf v}$. The joint correlation matrix $\boldsymbol C_{\mathbf a}$ across the A features ${\boldsymbol X}_{\mathbf a}$, and the combined A-V features $\boldsymbol J$ are given by: 
\begin{equation}
  \boldsymbol C_{a}= \tanh \left(\frac{{\boldsymbol X}_{ a}^\top{\boldsymbol W}_{ja}{\boldsymbol J}}{\sqrt d}\right)
\end{equation}
where ${\boldsymbol W}_{ja} \in\mathbb{R}^{d_a\times d} $ represents learnable weight matrix across the $\boldsymbol X_a$ and $\boldsymbol J$. 

Similarly, the cross-correlation matrices of other two modalities $ \boldsymbol C_{v}$ and $ \boldsymbol C_{t}$ are obtained as 
\begin{equation}
  \boldsymbol C_{v}= \tanh \left(\frac{{\boldsymbol X}_{ v}^\top{\boldsymbol W}_{jv}{\boldsymbol J}}{\sqrt d}\right)
\end{equation}
\begin{equation}
  \boldsymbol C_{t}= \tanh \left(\frac{{\boldsymbol X}_{ t}^\top{\boldsymbol W}_{jt}{\boldsymbol J}}{\sqrt d}\right)
\end{equation}
where ${\boldsymbol W}_{jv} \in\mathbb{R}^{d_v\times d}$, ${\boldsymbol W}_{jt} \in\mathbb{R}^{d_t\times d} $ represents learnable weight matrices for visual and text modalities respectively. 

The obtained joint cross-correlation matrices of the individual modalities are used to compute the attention weights, thereby capturing the semantic relevance of both across and within the same modalities. Higher correlation coefficient of the joint cross-correlation matrices denote higher semantic relevance pertinent to the intra- and inter-modal relationships of the corresponding feature vectors. Now the joint cross-correlation matrices are used to compute the attention maps of the individual modalities. For the audio modality, the joint correlation matrix $\boldsymbol C_{a}$ and the corresponding audio features ${\boldsymbol X}_{a}$ are combined using the learnable weight matrix $\boldsymbol W_{ca}$, followed by $ReLU$ activation function to compute the attention maps $\boldsymbol H_{a}$, which is given by 
\begin{equation}
\boldsymbol H_{a}=ReLU(\boldsymbol X_{a} \boldsymbol W_{ca} {\boldsymbol C}_a)
\end{equation}
where ${\boldsymbol W}_{\mathbf c \mathbf a} \in\mathbb{R}^{{K}\times {K}}$ denote learnable weight matrix for audio modality. Similarly the attention maps of visual and text modalities are obtained as 
\begin{equation}
\boldsymbol H_{v}=ReLU(\boldsymbol X_{v} \boldsymbol W_{cv} {\boldsymbol C}_v)
\end{equation}
\begin{equation}
\boldsymbol H_{t}=ReLU(\boldsymbol X_{t} \boldsymbol W_{ct} {\boldsymbol C}_t)
\end{equation}

where ${\boldsymbol W}_{cv} \in\mathbb{R}^{{K}\times {K}}, {\boldsymbol W}_{ct} \in\mathbb{R}^{{K}\times {K}}$ are the learnable weight matrices of visual and text modalities respectively.
Now the attention maps are used to compute the attended features of the individual modalities as: 
\begin{equation}
{\boldsymbol X}_{att, a} = \boldsymbol H_{a} \boldsymbol W_{ha} + \boldsymbol X_{a}
\end{equation}
\begin{equation}
{\boldsymbol X}_{att, v} = \boldsymbol H_{v} \boldsymbol W_{hv} + \boldsymbol X_{v}  
\end{equation}
\begin{equation}
{\boldsymbol X}_{att,t} = \boldsymbol H_{t} \boldsymbol W_{ht} + \boldsymbol X_{t}  
\end{equation}
where $\boldsymbol W_{ha} \in\mathbb{R}^{{K}\times {K}}$, $\boldsymbol W_{hv} \in\mathbb{R}^{{K}\times {K}}$, and $\boldsymbol W_{ht} \in\mathbb{R}^{{K}\times {K}}$ denote the learnable weight matrices for audio, visual and text modalities respectively. 

In order to obtain more refined feature representations, the attended features of each modality are again fed as input to the respective joint cross-modal attention module, which is given by  
\begin{equation}
{\boldsymbol X}_{att, a}^{(l)} = \boldsymbol H_{a}^{(l)} \boldsymbol W_{ha}^{(l)} + \boldsymbol X_{a}^{(l-1)}
\end{equation}
\begin{equation}
{\boldsymbol X}_{att,v}^{(l)} = \boldsymbol H_{v}^{(l)} \boldsymbol W_{hv}^{(l)} + \boldsymbol X_{v}^{(l-1)}  
\end{equation}
\begin{equation}
{\boldsymbol X}_{att,t}^{(l)} = \boldsymbol H_{t}^{(l)} \boldsymbol W_{ht}^{(l)} + \boldsymbol X_{t}^{(l-1)}  
\end{equation}

where $\boldsymbol W_{ha}^{(l)} \in\mathbb{R}^{{K}\times {K}}$, $\boldsymbol W_{ht}^{(l)} \in\mathbb{R}^{{K}\times {K}}$, and $\boldsymbol W_{ht}^{(l)} \in\mathbb{R}^{{K}\times {K}}$ denote the learnable weight matrices of audio, visual and text modalities respectively and $l$ refers to the recursive step. 

The attended features of the individual modalities after $l$ iterations are concatenated to obtain the multimodal feature representation ${\boldsymbol X}_{att}^{(l)}$, shown as: 
\begin{equation}
 {{\boldsymbol X}_{att}^{(l)} = [{\boldsymbol X}_{att,a}^{(l)} ; {\boldsymbol X}_{att,v}^{(l)} ; {\boldsymbol X}_{att,t}^{(l)}]}  
\end{equation}

Finally the multimodal feature representation ${\boldsymbol X}_{att}^{(l)}$ is fed to regression layers (Multi Layer Perceptron) for the final prediction of valence or arousal.

%The parameters of the proposed fusion model are optimized by minimizing this loss function.

\section{Experimental Setup}
\subsection{Dataset}
Affwild is the largest database in the field of affective computing, consisting of $298$ videos captured under extremely challenging conditions from YouTube videos \cite{Kollias}. The database is extended substantially to foster the development of robust models for estimating valence and arousal by introducing $260$ more videos, resulting in a total of $558$ videos with $1.4$M frames \cite{kollias2019expression}. The database has been further expanded by adding new videos in the subsequent series of ABAW challenges \cite{zafeiriou2017aff,kollias2020analysing,kollias2021analysing,kollias2022abaw,kollias2023abaw,kollias2023abaw2}. In the 6th ABAW challenge \cite{kollias20246th} for the valence-arousal track, the dataset is provided with $594$ videos of around $2,993,081$ frames from $584$ subjects. Sixteen of these videos display two subjects, both of which have been annotated. The final annotations are obtained by taking an average of annotations provided by four experts, using a joystick. The annotations for valence and arousal are provided continuously in the range of $[-1, 1]$. The dataset is partitioned into the training, validation, and test sets in a subject-independent manner to ensure that each subject appears exclusively in one partition. The partitioning resulted in $356$, $76$, and $162$ videos for train, validation, and test partitions respectively.

\subsection{Implementation Details}
\subsubsection{Preprocessing}
For the visual modality, we have used the cropped and aligned images provided by the challenge organizers \cite{kollias20246th}. For the missing faces in the video frames, we have considered black frames (i.e., zero pixels), and the video frames with no annotations of valence and arousal i.e., with annotations of $-5$ are excluded. The given video sequences are divided into sub-sequences of $300$ (i.e.,$K=300$) with a stride of $200$ frames and the facial images are resized to $48 \times 48$.  

For audio modality, the speech signals are extracted from the corresponding videos with a sampling rate of 16KHz. The log melspectrograms are then obtained using the preprocessing code provided by the Vggish repository\footnote{https://github.com/harritaylor/torchvggish}. To ensure that the audio modality is properly synchronized with the sub-sequences of other modalities, we have used a hop length of $1/fps$ of the raw videos to extract the spectrograms.      

For the text modality, the extracted speech signals from audio preprocessing are fed to the pretrained speech recognition model of Vosk toolkit\footnote{https://alphacephei.com/vosk/models/vosk-model-en-us-0.22.zip} to obtain the recognized words along with word-level timestamps. Next, a pretrained punctuation restoration and capitalization model\footnote{https://pypi.org/project/deepmultilingualpunctuation/} is used to restore the punctuations of the recognized words, which carries semantic information pertinent to emotional states. Now BERT features are extracted at word-level using a pre-trained BERT model\footnote{https://pypi.org/project/pytorch-pretrained-bert/}. The word-level features are computed by taking a summation of the last four layers of the BERT model \cite{sun2020multi}. The recognized words may usually span a larger time window of multiple frames. In order to synchronize the word-level BERT features of text modality with audio and visual modalities, the word-level text embedding is populated as per the timestamp of each word by reassigning the same word-level feature to all the frames within the time span of the corresponding word.

\subsubsection{Training Details}
For visual modality, random flipping, and random crop with a size of $40$ are used for data augmentation in training, while only center crop is used for validation. For audio and visual features, the input data is normalized in order to have a mean and standard deviation of $0.5$. For text modality, the BERT features are normalized to ensure the mean as $0$ and standard deviation as $1$. Adam optimizer is used with a weight decay of $0.001$ and batch size is set to be $12$. The models are trained separately for valence and arousal. The maximum number of epochs is set to be $100$ and early stopping is employed to avoid over-fitting. The hyperparameters of the initial learning rate and minimum learning rate are set to be $1e-5$ and $1e-8$ respectively. In our training strategy, we have deployed a warm-up scheme using $ReduceLROnPlateau$ scheduler with patience of $5$ and a factor of $0.1$ based on the CCC score of the validation partition. It has been shown that gradual training of the backbones of individual modalities along with the fusion model by gradually fine-tuning the layers of the backbones helps to improve the performance of the system \cite{10208757}. Therefore, we have deployed a similar strategy in our training framework, where three groups of layers for visual (Resnet-50) and audio (VGG) backbones are progressively selected for fine-tuning. Initially at epoch $0$, the first group is unfrozen and the learning rate is linearly warmed up to $1e-5$ within an epoch. Then repetitive warm-up is employed until epoch $5$, after which $ReduceLROnPlateau$ is used to update the learning rate. The learning rate is gradually dropped with a factor of $0.1$ until validation CCC does not improve over 5 consecutive epochs. After which the second group is unfrozen and the learning rate is reset to $1e-5$, followed by a warm-up scheme with $ReduceLROnPlateau$. The procedure is repeated till all the layers are fine-tuned for audio and visual backbones. Also, note that the best model state dictionary over prior epochs is loaded at the end of each epoch to mitigate the issues of over-fitting. To further control the problem of over-fitting, we have employed cross-validation with 6 folds, where the fold $0$ partition is the same as the original partition provided by the organizers \cite{kollias20246th}. The results obtained from the 6-fold cross-validation are shown in Table \ref{Cross validation}. In all these experiments, we have used 3 iterations in the fusion model (i.e., $l$=3).    
\begin{table}%[!b]
%\scriptsize
\renewcommand{\arraystretch}{1.25}
    \centering
    \caption{ CCC of the proposed approach on the six folds of cross-validation. Bold indicates the highest scores. Fold 0 is the official validation set.}
    \label{Cross validation}
    \begin{tabular}{|l|c|c|c|c|c|c||c|c|c|} %
	\hline
	 \textbf{Validation Set } & \textbf{Valence}  &  \textbf{Arousal} & \textbf{Mean}\\
	 \hline  \hline
 	 %\multicolumn{7}{|c|}{\textbf{Weakly-Supervised}} \\
	%\hline
	Fold 0  & 0.455 &  0.652 & 0.553 \\
	\hline
    Fold 1 & \textbf{0.585} & \textbf{0.674} & \textbf{0.629} \\
	\hline
	Fold 2 & 0.467 & 0.631 & 0.549 \\
	\hline
	Fold 3 & 0.536 & 0.647 & 0.591\\
	\hline
	Fold 4 & 0.432 & 0.629 & 0.526 \\
	\hline
   Fold 5 &  0.463 & 0.615 & 0.539 \\
	\hline
\end{tabular}
     %\vspace{-8mm}
\end{table}
\subsubsection{Loss Function}
The concordance correlation coefficient ($\rho_c$) is the widely-used evaluation metric in the literature for DER to measure the level of agreement between the predictions ($x$) and ground truth ($y$) annotations \cite{cite7}. Let $\mu_x$, and $\mu_y$ represent the mean of predictions and ground truth, respectively. Similarly, $\sigma_x^2$ and $\sigma_y^2$ denote the variance of predictions and ground truth, respectively, then $\rho_c$ between the predictions and ground truth can be obtained as:
\begin{equation}
\rho_c=\frac{2\sigma_{xy}^2}{\sigma_x^2+\sigma_y^2+(\mu_x-\mu_y)^2}
\end{equation}
where $\sigma_{xy}^2$ denotes the covariance between predictions and ground truth. %OK
%\subsection{Objective Function}
Though Mean Square Error (MSE) is the commonly used loss function for regression models, we use the CCC-based loss function as it is a standard loss function in the literature for DER \cite{Praveen_2022_CVPR, cite7}, which is given by 
\begin{equation}
\mathcal{L} = 1 - \rho_c     
\end{equation}
\begin{table*}
%\scriptsize
\setlength{\tabcolsep}{4pt}
\renewcommand{\arraystretch}{1.25}
    \centering
    \caption{ CCC of the proposed approach compared to state-of-the-art methods for multimodal fusion on the original Affwild2 validation set (fold 0). ${\star}$ indicates that results are presented with the implementation of our experimental setup. Highest scores are shown in bold.}
    \label{Comparison with state-of-the-art for Affwild2 validation}
    \begin{tabular}{|c|c|c|c|c|c|c|c|c|c|c|} 
	\hline
	\textbf{Method}& 
 \textbf{Type of } & \multicolumn{2}{|c|}{\textbf{Validation Set}} & \multicolumn{2}{|c|}{\textbf{Test Set}} \\
    %\midrule
    \cline{3-6}
     & \textbf{Fusion} & \textbf{Valence} & \textbf{Arousal} & \textbf{Valence} & \textbf{Arousal} \\
%    \midrule
\hline \hline
  %  Kuhnke et al. \cite{9320301}  & Resnet18 &  R(2plus1)D & - & Feature Concat & 0.493 & 0.604 \\
%	\hline
 %   Zhang et al. \cite{9607460}  & VGG &  Resnet50 & - & Leader-Follower & 0.405  & 0.645 \\
  %  \hline \hline
    Zhang et al. \cite{10208713} & Transformers & 0.464 & 0.640 & \textbf{0.648} & 0.625\\
  \hline
      Zhang et al. \cite{10208757} & Leader-Follower & 0.441 & 0.645 & 0.552 & 0.629\\
  \hline
  Zhou et al. \cite{10209026} & Transformers & 0.550 & \textbf{0.681} & 0.500 & \textbf{0.632}\\
  \hline
  Zhang et al. \cite{10208807} & Transformers & 0.554 & 0.659 & 0.523 & 0.545\\
  \hline
  Meng et al. \cite{Situ-RUCAIM3} & Transformers & 0.588 & 0.668 & 0.606 & 0.596\\
  \hline
  
%  Praveen et al. \cite{9667055}  & Resnet18 &  I3D & - & Cross-Attention  & 0.351  & 0.417 \\ 
%	\hline 
 Praveen et al \cite{Praveen_2022_CVPR} & JCA  & 0.663 & 0.584 & 0.374 & 0.363\\ 
	\hline
 Praveen et al \cite{10095234} & RJCA  & \textbf{0.703} & 0.623 & 0.467 & 0.405\\ \hline
 	Praveen et al \cite{10095234}$^{\star}$ & RJCA  & 0.443 & 0.639 & 0.537  & 0.576\\ 
	\hline
	RJCMA (Ours) & RJCMA  & 0.455 & 0.652 & 0.542 & 0.619 \\ 
	\hline
\end{tabular}
     %\vspace{-8mm}
\end{table*}
\begin{table}%[!b]
%\scriptsize
\renewcommand{\arraystretch}{1.25}
    \centering
    \caption{ CCC of the proposed fusion model by varying the number of recursions. Bold indicates the highest scores. Fold 1 is used for experiments with multiple recursions.}
    \label{Ablation Study}
    \begin{tabular}{|l|c|c|c|c|c|c||c|c|c|} %
	\hline
	 \textbf{Num. of recursions ($l$)} & \textbf{Valence}  &  \textbf{Arousal} & \textbf{Mean}\\
	 \hline  \hline
 	 %\multicolumn{7}{|c|}{\textbf{Weakly-Supervised}} \\
	%\hline
	$l$ = 1  & 0.568 & 0.656 & 0.607 \\
	\hline
    $l$ = 2 & 0.576 & 0.669 & 0.618 \\
	\hline
	$l$ = 3 & \textbf{0.585} & \textbf{0.674} & \textbf{0.629} \\
	\hline
	$l$ = 4 & 0.579 & 0.652 & 0.615 \\
	\hline
\end{tabular}
     %\vspace{-8mm}
\end{table}

\section{Results and Discussion}
\begin{table*}%[!b]
%\scriptsize
\renewcommand{\arraystretch}{1.25}
    \centering
    \caption{ CCC of the proposed approach on Affwild2 test set compared to other methods submitted to 6th ABAW competition. Bold indicates best results of our approach on the test set.${*}$ denote citations are not available.} 
    \label{Comparison with state-of-the-art for Affwild2 test}
    \begin{tabular}{|l|c|c|c|c|c|c||c|c|c|} %
	\hline
	 \textbf{Method } & \textbf{Modalities} & \textbf{Ensemble}
	 & \textbf{Valence}  &  \textbf{Arousal} & \textbf{Mean}\\
	 \hline  \hline
 	 %\multicolumn{7}{|c|}{\textbf{Weakly-Supervised}} \\
	%\hline
	Netease Fuxi AI Lab \cite{zhang2024affective}  & Audio, Visual  & Yes &   0.6873 & 0.6569 & 0.6721 \\
	\hline
    RJCMA (Ours) & Audio, Visual, Text  & No & \textbf{0.5418} & \textbf{0.6196} &  \textbf{0.5807}\\
	\hline
	CtyunAI \cite{zhou2024boosting} & Visual & No & 0.5223 & 0.6057 & 0.5640 \\
	\hline
	SUN-CE \cite{dresvyanskiy2024sun} & Audio, Visual & Yes &0.5355 & 0.5861 &  0.5608\\
	\hline
	USTC-IAT-United \cite{yu2024multimodal} & Audio, Visual & No & 0.5208 & 0.5748 & 0.5478 \\
	\hline
   HSEmotion \cite{savchenko2024hsemotion} & Visual & No & 0.4925 & 0.5461 &  0.5193 \\
	\hline
    KBS-DGU$^{*}$ & Audio, Visual & No &  0.4836 & 0.5318 &  0.5077 \\
	\hline
    ETS-LIVIA \cite{waligora2024joint} & Audio, Visual & No & 0.4198 & 0.4669 &  0.4434 \\
	\hline
    CAS-MAIS$^{*}$ & Audio, Visual, Text & No & 0.4245 & 0.3414 &  0.3830 \\
	\hline
    IMLAB \cite{min2024emotion} & Visual & No & 0.2912 & 0.2456 &  00.2684 \\
	\hline
    Baseline \cite{kollias20246th} & Visual & No & 0.2110 & 0.1910 &  0.2010 \\
	\hline
\end{tabular}
     %\vspace{-8mm}
\end{table*}
\subsection{Ablation Study}
In order to understand the impact of the recursive mechanism, we have conducted a series of experiments by varying the number of recursions as shown in Table \ref{Ablation Study}. First, we did an experiment with a single recursion, which is the same as joint cross attention \cite{Praveen_2022_CVPR} for audio, visual, and text modalities. Now we executed multiple experiments by slowly increasing the number of recursions and found that the performance of the system gradually increases with multiple recursions. This shows that the recursive mechanism helps in obtaining more robust feature representations by progressively refining the features. We have achieved the best results at 3 iterations, beyond that the performance of the system declines. We hypothesize that this can be attributed to the fact that though recursive fusion initially helps to improve the performance, it may result in over-fitting with more iterations, resulting in a performance decline on the validation set. A similar trend of performance improvement with multiple recursions is also observed for other audio-visual tasks such as person verification \cite{praveen2024audio} and event localization \cite{9423042}, thereby underscoring our hypothesis. %It is worth mentioning that the results are reported with an average of three runs for statistical stability.      

\subsection{Comparison to State-of-the-art}
Most of the approaches evaluated on the Affwild2 dataset have been submitted to previous ABAW challenges. Therefore, we have compared the performance of the proposed approach with some of the relevant state-of-the-art models of previous ABAW challenges as shown in Table \ref{Comparison with state-of-the-art for Affwild2 validation}. Several approaches explored ensemble-based methods using multiple encoders for each modality, followed by transformers to encode the concatenated multimodal feature representation \cite{10209026,10208807,Situ-RUCAIM3}. By exploiting multiple backbones and large-scale training with external datasets, Meng et al. \cite{Situ-RUCAIM3} significantly improved the test set performance on both valence and arousal. Similarly, Zhou et al. \cite{10209026} also explored multiple backbones and showed better improvement in the arousal performance of the test set than \cite{Situ-RUCAIM3}. Even though ensemble-based methods show better performance on test set, it is often cumbersome and computationally expensive. Zhang et al. \cite{10208713} showed that exploring Masked Auto-Encoders (MAE) is a promising line of research to achieve better generalization and consistently improved performance in both valence and arousal. Zhang et al. \cite{10208757} explored the leader-follower attention model for multimodal fusion, where audio and text modalities are leveraged to attend to the visual modality, and showed good performance on the test set without the need for multiple backbones for each modality.

Praveen et al. \cite{Praveen_2022_CVPR} proposed Joint Cross Attention (JCA) by introducing a joint feature representation in the cross-attention framework and showed significant improvement in the validation set, especially for valence. They further improved the performance of their approach by deploying a recursive mechanism and LSTMs for the temporal modeling of individual modalities and multimodal feature representations \cite{10095234}. However, they do not seem to have a better generalization ability as they fail to show improvement in the test set. We hypothesize that this may be due to naive audio and visual backbones, fine-tuned on Affwild2 dataset. Therefore, to have a fair comparison with \cite{10095234}, we have reimplemented their fusion model by replacing their audio and visual backbones with Vggish \cite{hershey2017cnn} and Resnet-50 pre-trained on MSCELEB-1M and FER+ \cite{10208757}, followed by TCNs. By replacing the visual and audio backbones of \cite{10095234} and progressive fine-tuning of the backbones similar to that of \cite{10208757}, the problem of over-fitting has been mitigated and showed improvement in test set performance also. We can observe that performance has been improved in both validation and test sets by introducing text modality into the RJCA framework \cite{10095234}, especially for arousal. %Since there is only marginal improvement with the text modality, we expect that the performance of the system can be further improved by leveraging advanced text encoders. %It is worth mentioning that the results are reported with an average of three runs for statistical stability. 
Note that even though the performance of the valence for \cite{10208713,10208757} as well as our approach on the official validation set is lower, they achieved better performance on other folds of cross-validation, thus improving the performance of the test set. 
%minimal improvement in fusion performance over uni-modal performances. Though they have shown significant performance for arousal than valence, it is mostly attributed to the V backbone. Similar to that of \cite{9607460}, we also follow the same backbones for the audio, visual, and text modalities. BY deploying the cross-modal attention in a recursive fashion, we are able to achieve better results than that of the relevant methods on the validation set of Affwid2. Even with vanilla cross-attentional fusion \cite{9667055}, the fusion performance for valence has been improved better than that of \cite{9607460} and \cite{9320301}. By deploying joint representation into the cross attentional fusion model, the fusion performance of valence has been significantly improved further. In this work, we further extended our previous work \cite{Praveen_2022_CVPR} by introducing text modality and recursive fusion to further improve the performance of the system. 

\subsection{Results on Test Set}
%\subsection{Visual Results}
Multimodal learning is gaining a lot of attention for affective behavior analysis as most of the methods submitted for the valence-arousal challenge of the 6th ABAW competition \cite{kollias20246th} employed multimodal fusion \cite{zhang2024affective,dresvyanskiy2024sun,yu2024multimodal,waligora2024joint}. Some of these methods explored ensemble-based fusion for better generalization ability \cite{zhang2024affective,dresvyanskiy2024sun}. Another widely explored strategy to improve the test set performance is to exploit pre-training with multiple large-scale datasets using Masked Auto-Encoders (MAE) \cite{zhang2024affective,zhou2024boosting} or multiple backbones for each modality \cite{dresvyanskiy2024sun,savchenko2024hsemotion,yu2024multimodal}. Netease Fuxi AI Lab \cite{zhang2024affective} used MAEs pretrained with 5 external data sets of around 262M images and Vggish model \cite{hershey2017cnn} for the audio modality, followed by ensemble-based fusion, and showed a significant improvement for both valence and arousal, achieving first place in the challenge. Similarly, CtyunAI \cite{zhou2024boosting} also used MAE for visual modality pre-trained with 4 external datasets, followed by TCNs for temporal modeling, and achieved very good performance for arousal than valence. Sun-CE \cite{dresvyanskiy2024sun} explored multiple backbones for both audio and visual modalities, followed by ensemble-based fusion, while USTC-IAT-United \cite{yu2024multimodal} used multiple backbones only for the audio modality and achieved decent performance for both valence and arousal. KBS-DGU and ETS-LIVIA \cite{waligora2024joint} exploited combining features from multiple fusion models of self-attention and cross-attention models for improving the performance of audio-visual fusion. KBS-DGU explored both self-attention and cross-attention models, while ETS-LIVIA \cite{waligora2024joint} used joint feature representation with cross-attention models across audio and visual modalities. Unlike other approaches, we have explored the prospect of simultaneously capturing the intra- and inter-modal relationships across audio, visual, and text modalities recursively and achieved very good performance for both valence and arousal without using multiple backbones or pre-training with large-scale external datasets. Therefore, the performance of our approach can be solely attributed to the robustness of the sophisticated fusion model, providing a cost-effective solution for affective behavior analysis in-the-wild. % using multimodal fusion. %Since there is only marginal improvement with the inclusion of text modality, we also submitted the test set predictions from the best performing models on the six folds for both audio-visual and audio-visual-text systems. 

\section{Conclusion}
In this work, we showed that effectively capturing the synergic relationships pertinent to the intra- and inter-modal characteristics among the audio, visual, and text modalities can significantly improve the performance of the system. By introducing the joint representation in the cross-attentional framework, we can simultaneously capture both intra- and inter-modal relationships across the audio, visual, and text modalities. The performance of the system is further enhanced using recursive fusion by progressively refining the features to obtain robust multimodal feature representations. Experimental results in the challenging Affwild2 dataset indicate that the proposed model can achieve a better multimodal fusion performance, outperforming most methods for both valence and arousal. The performance of the system can also be improved by leveraging advanced text encoders and sophisticated backbones for the individual modalities.  

\noindent \textbf{Acknowledgements } The authors wish to acknowledge the funding from the Government of Canada’s New Frontiers in Research Fund (NFRF) through grant NFRFR-2021-00338 and Ministry of Economy and Innovation (MEI) of the Government of Quebec for the continued support.
{
    \small
    \bibliographystyle{unsrt} 
    \bibliography{main}
}


\end{document}